\documentclass[10pt,twocolumn,letterpaper]{article}

\usepackage{cvpr}

\makeatletter
\AtBeginDocument{%
  \@ifpackageloaded{lineno}{\setlength{\linenumbersep}{1.45cm}}{}%
}
\makeatother

\usepackage{amsmath}
\usepackage{amssymb}
\usepackage{bm}
\usepackage{booktabs}
\usepackage{makecell}
\usepackage{multirow}
\usepackage{graphicx}
\usepackage{placeins}
\usepackage{adjustbox}
\usepackage{float} 
\usepackage{dblfloatfix} 

\graphicspath{{./}{figure/}}

\definecolor{cvprblue}{rgb}{0.21,0.49,0.74}
\iftoggle{cvprfinal}{%
  \usepackage[pagebackref,breaklinks,colorlinks,allcolors=cvprblue]{hyperref}%
}{%
  \usepackage[breaklinks,colorlinks,allcolors=cvprblue]{hyperref}%
}

\def\paperID{}
\def\confName{CVPR}
\def\confYear{2026}

\title{Semantic-Aware Guided Drone Exploration for Language-Conditioned 3D Indoor Mapping}

\author{Nitin Vegesna and Avideh Zakhor\\
Department of Electrical Engineering and Computer Sciences\\
University of California, Berkeley\\
Berkeley, CA, USA\\
{\tt\small \{nitinvegesna, avz\}@berkeley.edu}
}

\begin{document}
\maketitle
\begin{abstract}
We present Semantic-Aware Guided Exploration, SAGE, a system for open-vocabulary exploration in unknown 3D indoor environments that preserves coverage-oriented behavior while allowing semantic cues to reprioritize frontier selection.
Building on the FALCON volumetric explorer, SAGE integrates Contrastive Language-Image Pre-training (CLIP) via four key components: object-centric embedding storage, a temporal cache that projects recent observations onto the free--unknown boundary, object frontiers for high-similarity detections, and a unified semantic--geometric planning cost.
This cost function bounds semantic reweighting influence, ensuring frontiers are prioritized without sacrificing total coverage.
In Matterport3D-based simulations, SAGE outperforms FALCON and a semantic-only ablation in object discovery across map--query pairs.
Compared to Finding Things in the Unknown (FTU)~\cite{Papatheodorou_ICRA2023}, SAGE completes exploration $9.0$ to $25.9$ times faster across the nine shared map--query pairs, achieving a mean speedup of $13.7$. Furthermore, SAGE achieves substantially higher volumetric throughput than FTU.
Finally, we deploy SAGE in five real-world flights in two environments on a Modal AI Starling~2 quadrotor with onboard sensing and planning, and offboard CLIP inference.
Comparing SAGE and FALCON, we find that while FALCON results in faster exploration and shorter mapping trajectories, SAGE outperforms FALCON in terms of object discovery.
\end{abstract}
\section{Introduction}
\label{sec:intro}

Embodied agents that map and act in 3D indoor scenes increasingly rely on language, enabling users to specify targets with open-vocabulary phrases rather than fixed category identifiers.
Work on 3D-scene large language models, vision--language navigation, and vision--language--action models assumes that robots can ground words in perception and use that signal for decision-making.

In this work, we propose a novel language-conditioned volumetric exploration method.
A mobile platform with an RGB--D sensor operates in an initially unknown indoor environment to build a metric occupancy representation through active viewpoint selection.
The user provides an open-vocabulary text query naming a category or object of interest, such as \texttt{chair}.
The task is to jointly (i) drive the map toward full volumetric coverage---systematically reducing unknown space until exploration is complete in the frontier-based sense used in geometric planners---and (ii) discover all query-relevant object instances by approaching them within a defined proximity threshold.
The objective is to satisfy both goals as quickly and efficiently as possible, recognizing that coverage and object discovery often exert competing pressures on the robot's trajectory.

Frontier-based information seeking draws the robot toward unknown boundaries, while our CLIP-based similarity encourages viewpoints where the camera sees query-matching objects. Such objects can lie far from the nearest frontier or deep within partially explored space.
Current methods typically excel at only one task. Purely geometric exploration relies on chance to discover objects along a coverage path. Conversely, relying solely on semantics often results in inefficient movement. Classical frontier exploration~\cite{yamauchi1997frontier} and modern geometric planners such as FALCON~\cite{zhang2024falcon} optimize for coverage and path cost, but treat object discovery as an incidental byproduct.
In practice, FALCON uses a drone's depth stream to maintain a 3D occupancy map. It selects viewpoints that allow the robot to sweep through a space while keeping paths short. In contrast, object-centric semantic explorers~\cite{Papatheodorou_ICRA2023,barbas2025findanything,kim2025raven} prioritize relevant objects; however, these methods may take much longer to achieve full coverage or may fail to explore the environment systematically.
Bridging language, 3D perception, and motion requires a planner that fuses and balances geometric and language-conditioned cues during each replanning cycle. We achieve this by introducing semantic scale factors for the geometric cost functions, allowing semantics to reorder frontier priorities without overriding the underlying incentives for geometric coverage.

We propose Semantic-Aware Guided Exploration (SAGE), building on FALCON's existing pipeline for volumetric 3D mapping, unexplored-boundary detection, coverage tour planning, and smooth trajectory generation. SAGE redefines the way candidate viewpoints are scored and ordered to incorporate language via four primary mechanisms.
First, SAGE attaches CLIP embeddings to occupied surfaces and aggregates them into compact object-level memory so that the planner can reason about which parts of the map are likely to contain relevant objects.
Second, a short-lived temporal cache projects recent semantic observations toward the free--unknown boundary, allowing frontiers to receive similarity scores before the robot commits to visiting them.
Third, object frontiers add dedicated viewpoints that look directly at high-similarity objects and encourage close-up views.
Finally, a unified semantic--geometric cost replaces pure path length so that semantic similarity biases both the inter-cell tour and frontier-level viewpoint ordering.

This paper is organized as follows: Section~\ref{sec:related} reviews related work, Section~\ref{sec:method} describes the SAGE method, Section~\ref{sec:experiments} presents simulation and hardware experiments, and Section~\ref{sec:conclusion} concludes with key findings and limitations.

\section{Related Work}
\label{sec:related}

\paragraph{Geometric exploration.}
Frontier-based methods maximize coverage of unknown space~\cite{yamauchi1997frontier}, while information-theoretic variants maximize expected information gain~\cite{thrun2005probabilistic}.
Hierarchical and sampling-based 3D planners reduce computational load while maintaining systematic expansion~\cite{zhou2021fuel,dong2026eden}.
Representative 3D aerial exploration planners include receding-horizon next-best-view and information-driven frontier methods~\cite{bircher2016receding,dai2020fast,batinovic2021multiresolution,dornhege2011frontier}.
FALCON~\cite{zhang2024falcon} couples a coarse Traveling Salesman Problem (TSP) over cell-center waypoints with a local Sequential Ordering Problem (SOP) over frontier viewpoints, achieving strong 3D coverage using Micro Aerial Vehicles (MAVs) with real-time performance.
SAGE follows this stack so that any change in object-discovery behavior can be attributed to semantic guidance rather than alterations to the underlying geometric planning.

\paragraph{Vision--language cues in robotics.}
CLIP~\cite{radford2021clip} and successor vision--language pre-training models such as BLIP~\cite{li2022blip} and SigLIP~\cite{zhai2023sigmoid} enable open-vocabulary similarity between images and text, supporting query-driven navigation, inspection, and retrieval.
Closed-vocabulary detectors~\cite{ren2015fasterrcnn,redmon2016yolo} remain common in robotics but require curated label sets that do not match free-form user language.
Recent 3D and embodied systems combine transformers with point clouds or meshes for scene captioning, question-answering, and instruction-following~\cite{wang2022spacap3d,azuma2022scanqa,ye2021tvcg3dqa,goetting2025vlmnav}. Our focus is complementary; we use a classical volumetric planner whose edge costs are shaped by CLIP, ensuring the method remains compatible with small onboard computers once embeddings are available.

\paragraph{Object-centric and semantic exploration.}
Systems such as RAVEN~\cite{kim2025raven}, FindAnything~\cite{barbas2025findanything}, and Finding Things in the Unknown (FTU)~\cite{Papatheodorou_ICRA2023} emphasize semantic maps and object-directed behavior.
Recent work on semantic map representations and goal-directed semantic exploration further motivates this direction~\cite{raychaudhuri2025semantic,huang23vlmaps,conceptfusion2023,nagami2025vista,chaplot2020object,ye2021objectnav,ginting2024seek}.
FTU couples instance segmentation and semantic submaps in Habitat~\cite{savva2019habitat}; such pipelines typically rely on closed-set instance segmentation~\cite{he2017maskrcnn} and therefore remain tied to a fixed category vocabulary.
We compare SAGE to FTU on shared Matterport layouts, using each system's native stack and simulator.

\paragraph{Reconstruction-centric exploration.}
NeRF-guided and 3D Gaussian splatting--guided methods, such as~\cite{tao2024rtguide,kerbl2023gaussian}, optimize view quality for photometric reconstruction.
These methods excel when the goal is a high-fidelity model for rendering; however, by default, they do not optimize language-specified object discovery under a completeness-style stopping rule. They often assume sensor noise models that differ from those encountered on low-cost MAV RGB--D platforms.

\section{Method}
\label{sec:method}

We present SAGE as an extension of FALCON's volumetric frontier planner.
In Sections~\ref{sec:semantic_memory} and~\ref{sec:temporal_cache} we define $\mathbf{e}_R$ and~$\mathbf{e}_F$, the pooled cell-center and frontier CLIP embeddings, respectively.
In Section~\ref{sec:unified_cost} we use them to form the TSP and SOP semantic--geometric edge costs, so exploration can combine language similarity with obstacle-aware path length.

\subsection{Background}
\begin{figure}[t]
  \centering
  \includegraphics[width=\linewidth,keepaspectratio]{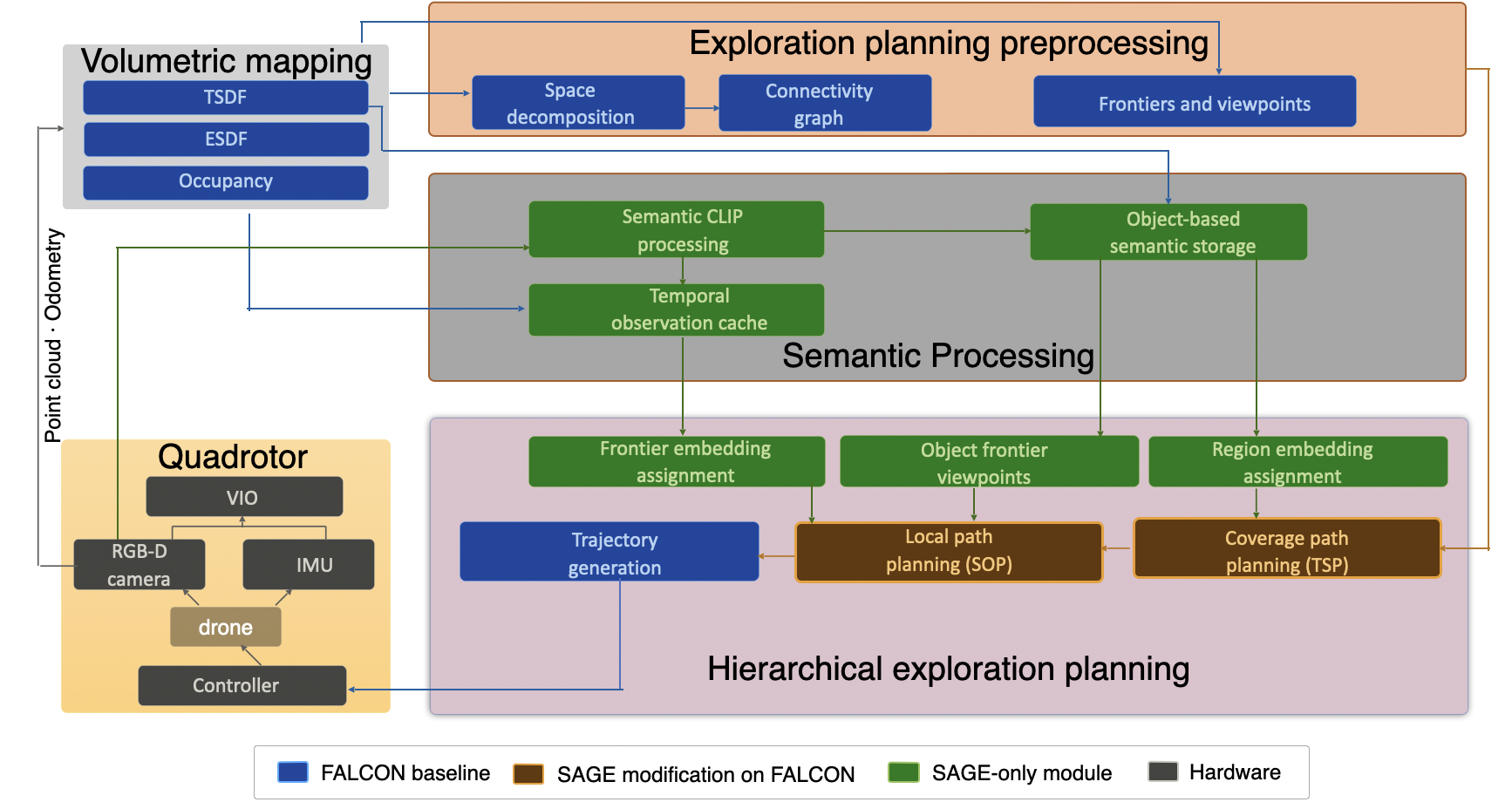}
  \caption{SAGE pipeline. Blue indicates blocks from FALCON. Green shows blocks added to SAGE and brown shows modified FALCON blocks.}
  \label{fig:pipeline}
\end{figure}

At a high level, SAGE layers language-grounded semantics onto FALCON's hierarchical planning through semantic processing.
It remembers where observations match the text query, propagates that signal toward the free--unknown boundary, and adjusts viewpoint order while keeping FALCON's coverage-oriented backbone.\newline
\hspace*{\parindent}Figure~\ref{fig:pipeline} shows the overall block diagram of SAGE. Relative to FALCON, the same three phases---volumetric mapping, exploration-planning preprocessing, and hierarchical exploration planning---are augmented by a fourth phase, semantic processing, that contains the CLIP-backed perception and memory blocks feeding the planner. Within each phase, blue denotes unmodified FALCON blocks, green indicates new SAGE modules, and brown corresponds to FALCON modules modified in SAGE.
The baseline blue blocks---volumetric mapping, space decomposition, connectivity graph and trajectory generation---are the same as in FALCON. Frontiers and viewpoint sampling, coverage path planning, and local path planning are extended in SAGE and shown in the brown blocks. The green blocks---semantic CLIP processing, object-based semantic storage, temporal observation cache, frontier embedding assignment, region embedding assignment, and object frontier generation---are new SAGE modules.
The overall objective is to favor regions and frontier viewpoints that are semantically similar to the text query without sacrificing the systematic coverage and path efficiency of the underlying geometric planner.
We achieve this by adding semantics to the inputs of the cost functions in the coverage and local path planning blocks shown in brown in Figure~\ref{fig:pipeline}.

\subsection{Semantic memory on the TSDF}
\label{sec:semantic_memory}
\begin{figure}[t]
  \centering
  \includegraphics[width=\linewidth]{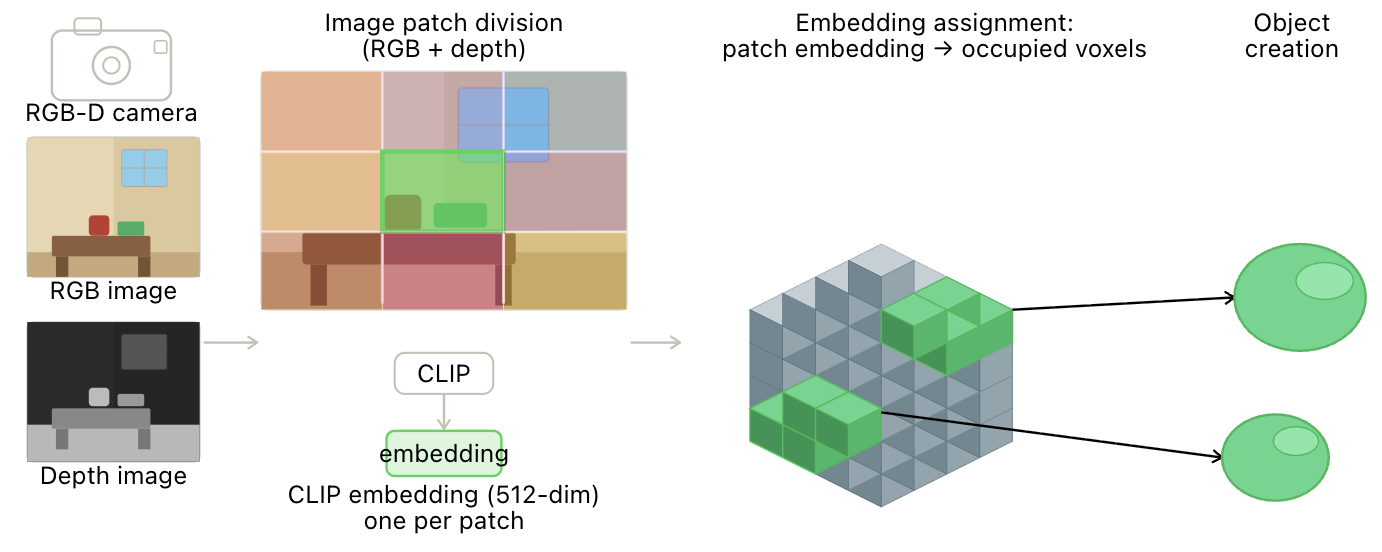}
  \caption{Object-embedding creation in semantic memory. RGB--D patches are encoded by CLIP, patch embeddings are assigned to occupied TSDF voxels, and object embeddings are created or updated from components defined on those voxels.}
  \label{fig:semantic_processing_pipeline}
\end{figure}

In this section, we describe the semantic CLIP processing and object-based semantic storage blocks in Figure~\ref{fig:pipeline}. Our strategy is to keep language-grounded object instances on observed surfaces without a dense per-voxel CLIP map.
Figure~\ref{fig:semantic_processing_pipeline} summarizes the patch-to-object instance flow.
Within the ViT-B/32~\cite{dosovitskiy2021image} image encoder in CLIP~\cite{radford2021clip}, the image is tiled into a regular grid of non-overlapping square patches, and each patch is associated with a separate 512-D embedding tied to a localized image region rather than a single global description of the full image.
For each patch, we convert the corresponding depth map, together with intrinsics and pose, into a point cloud.
We take the occupied TSDF voxels that those points fall into and partition them into connected components on the grid, so a single patch can yield several components when it views spatially separated surfaces.
Each patch's CLIP cosine similarity to the text is compared to the running maximum of patch--query similarity on the trajectory. 
Semantic memory is built from patches whose similarity is at least half of that maximum, so unrelated views do not populate storage. Components that pass this gate become object instances.\newline
\hspace*{\parindent}The process of object creation and update happens continuously as patches for new depth frames are captured and processed. 
If object voxels from the patches of the current RGB--D frame already formed an object embedding due to patches in previous frames, the stored object embedding is updated.
Otherwise, we create a new object and its corresponding embedding. 
Specifically, for each newly created component, let $\mathbf{p}$ be its $\ell_2$-normalized patch-CLIP embedding.
To associate a component with a stored object instance, the component must intersect the instance's bounding box, have a minimum vertical overlap that limits spurious association with large horizontal structures such as the floor, and have cosine similarity between~$\mathbf{p}$ and the stored instance embedding~$\mathbf{e}$ above a threshold.
If a stored object instance meets these conditions, we fuse the two embeddings with an exponential moving average of step size $\alpha \in (0,1)$:
\begin{equation}
\mathbf{e}\leftarrow
\dfrac{(1-\alpha)\mathbf{e}+\alpha\mathbf{p}}{\left\lVert(1-\alpha)\mathbf{e}+\alpha\mathbf{p}\right\rVert_2}
\end{equation}
We decrease $\alpha$ as the instance's occupied voxel count grows so that large objects resist noisy drift.
If no stored object instance meets the conditions, we create a new one and set $\mathbf{e}\leftarrow\mathbf{p}$.
\newline
\hspace*{\parindent}In FALCON, TSP plans a tour over cell-center waypoints in free and unknown subregions. We use these object instances to provide an embedding~$\mathbf{e}_R$ for the subregion inputs to the TSP.
We set~$\mathbf{e}_R=\mathbf{0}$ for unknown-subregion centers.
At each free-subregion center, we form~$\mathbf{e}_R$ by distance-weighted averaging of object embeddings that fall within~$r$ of that center.
Specifically, let $i$ index all object instances within range~$r$ of the free-subregion center; let $\mathbf{e}_i$ be instance~$i$'s CLIP embedding and~$d_i$ the Euclidean distance from that center to instance~$i$.
Define weights $w_i=1/(d_i+\epsilon)$ for a small fixed $\epsilon>0$.
If no object instance is in range, we set $\mathbf{e}_R=\mathbf{0}$.
Otherwise, we set:
\begin{equation}
  \begin{aligned}
    \mathbf{v} &= \frac{\sum_i w_i \mathbf{e}_i}{\sum_i w_i} \\
    \mathbf{e}_R &= \mathbf{v}/\lVert\mathbf{v}\rVert_{\ell_2}
  \end{aligned}
\end{equation}
Storing only object embeddings keeps typical memory footprint around $\sim$10~MB rather than $\sim$1--2~GB for a dense per-voxel field.
We use these object instances to create object frontiers in Section~\ref{sec:object_frontiers}, and we use $\mathbf{e}_R$ in TSP's cost formulation in Section~\ref{sec:unified_cost}.

\subsection{Temporal cache and frontier embeddings}\label{sec:temporal_cache}
In this section, we describe how we build and query a short-lived temporal cache to form the frontier embedding~$\mathbf{e}_F$ used in the SOP cost in Section~\ref{sec:unified_cost}. 
This is shown as the green blocks named ``temporal observation cache'' and ``frontier embedding assignment'' in Figure~\ref{fig:pipeline}. FALCON runs a sequence of replanning cycles. In each cycle, the TSP and SOP solvers run to select the next viewpoint and generate a trajectory.
In SAGE, we modify the cost function used within SOP by assigning the frontier embedding $\mathbf{e}_F$ via a temporal cache that is reset immediately after SOP is completed in a given cycle.

\begin{figure}[t]
  \centering
  \includegraphics[width=0.82\linewidth,keepaspectratio]{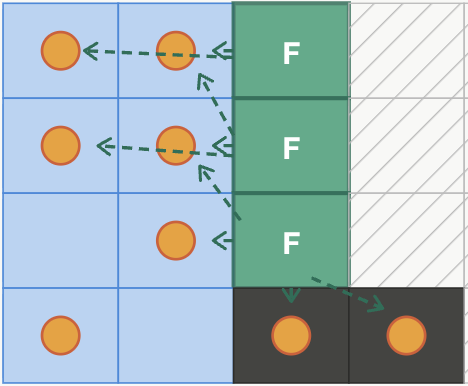}
  \caption{Illustration of temporal-cache queries at a frontier cluster. Green cells are frontier-cluster boundary voxels at the free--unknown interface. Blue indicates free voxels. Black stands for occupied voxels. Orange indicates cached entries in voxels. Dashed arrows indicate cache lookups issued while assembling the frontier embedding.}
  \label{fig:tcache}
\end{figure}

For every depth pixel in every incoming RGB--D frame, we trace a line-of-sight ray from the camera to the 3D point for that pixel through the TSDF voxel grid and write that pixel's CLIP patch embedding into every voxel, including free ones, that the ray passes through.
Figure~\ref{fig:tcache} illustrates cache queries at a frontier cluster.
Green marks the free--unknown boundary; blue and black voxels mark free and occupied space, respectively.
The orange circles mark all voxels included in the temporal cache, and dashed arrows show neighbor lookups.
For each frontier voxel we use the cache if it has an entry; otherwise we fuse from neighboring cached voxels, as shown in Figure~\ref{fig:tcache}.
We average over all voxels in the frontier cluster to form its embedding~$\mathbf{e}_F$ and set~$c_F$ to the fraction of frontier voxels where the cache lookup had nonzero support.

When a frontier persists across replans, we may merge its previous and current~$\mathbf{e}_F$ and update~$c_F$ from the merged pool.
If no cache support is available, we fall back to object embeddings along the ray from the robot pose to the frontier viewpoint and pool them into~$\mathbf{e}_F$ with~$c_F=0.5$. This ray is a cheap proxy for line-of-sight evidence between the current pose and the target viewpoint.
If that semantic cue is also unavailable for a frontier cluster, we set both $\mathbf{e}_F$ and $c_F$ to zero.

\subsection{Object frontiers for SOP}\label{sec:object_frontiers}
The input to the SOP module in FALCON is frontier viewpoints.
In SAGE, we add a new input to local SOP planning, namely object-frontier viewpoints, as shown in Figure~\ref{fig:pipeline}.
We now explain how we choose such viewpoints for a given object instance and how many object instances are considered in each replan.
Recall from Section~\ref{sec:semantic_memory} that each RGB--D frame builds up persistent object instances in semantic memory.
We consider object instances within range~$r_{\mathrm{obj}}$ and with instance--query CLIP cosine at least half the running max of patch--query similarity, per Section~\ref{sec:semantic_memory}.
From the qualifying set we take up to the three most similar instances per subregion, favoring pairwise separation.
We cast three horizontal rays from the free-subregion center at equidistant heights~$z$ so as to span the vertical extent of the object.
Each ray traverses the $xy$-plane toward the projected object center at voxel resolution while remaining in the same free subregion.
Among the three rays, we pass the collision-free endpoint nearest to the object to the SOP.
We repeat the same process for the ``next'' free subregion.
This is possible because the TSP creates a sequential order over the free and unknown subregions that the MAV is required to traverse.
These object-frontier candidates for the current and next subregion are added to the set evaluated by SOP in each replanning cycle, thereby shaping which viewpoint is selected next.

\subsection{Unified semantic--geometric cost}
\label{sec:unified_cost}
In this section, we modify FALCON's purely geometric cost functions in SOP and TSP to
incorporate semantics using $\mathbf{e}_F$ and $\mathbf{e}_R$ from Sections~\ref{sec:temporal_cache} and~\ref{sec:semantic_memory}, respectively.
This corresponds to modifying the ``coverage path planning'' and ``local path planning'' brown blocks in Figure~\ref{fig:pipeline}.
The SOP cost matrix records geometric path lengths between every pair of frontier viewpoints and from the current pose to each viewpoint.
Let $d^{\mathrm{SOP}}_{\mathrm{geo}}(i,j)$ denote the geometric path cost in meters from either frontier viewpoint~$i$ or the drone's pose to frontier viewpoint~$j$.
In SAGE, we scale $d^{\mathrm{SOP}}_{\mathrm{geo}}(i,j)$ by $m_{\mathrm{SOP}}(j)$ as follows:
\begin{equation}
  d^{\mathrm{SOP}}_{\mathrm{mod}}(i,j) = d^{\mathrm{SOP}}_{\mathrm{geo}}(i,j)\, m_{\mathrm{SOP}}(j)
  \label{eq:sop_mod_cost}
\end{equation}
The semantic factor $m_{\mathrm{SOP}}(j)$ in Equation~\eqref{eq:sop_mod_cost} is applied only to regular frontier viewpoints, and not to the object-frontier viewpoints described in Section~\ref{sec:object_frontiers}. This is because those candidates are already similarity-gated during generation and additional scaling would double-count semantic bias and over-prioritize object-frontier visits.
The scale factor $m_{\mathrm{SOP}}(j)$ in Equation~\eqref{eq:sop_mod_cost} is a function of pooling confidence~$c_F^j$ and similarity~$s_F^j$, where~$s_F^j$ is the cosine similarity between the text query and the pooled frontier embedding~$\mathbf{e}_F$ at~$j$, normalized by the running maximum of patch--query CLIP similarity.
Specifically, for each destination~$j$ we set
\begin{equation}
  U_{\mathrm{frontier}}\bigl(s_F^j,c_F^j\bigr)=\frac{4.0\,s_F^j+0.5\,c_F^j}{4.5}
  \label{eq:utility_function}
\end{equation}
\begin{equation}
  m_{\mathrm{SOP}}(j)=\max\!\left(0.2,\,1-U_{\mathrm{frontier}}\bigl(s_F^j,c_F^j\bigr)\right)
  \label{eq:sop_scale_factor}
\end{equation}

Similarly, let $d^{\mathrm{TSP}}_{\mathrm{geo}}(i,j)$ denote the geometric path cost from cell center~$i$ to cell center~$j$.
In SAGE, we scale $d^{\mathrm{TSP}}_{\mathrm{geo}}(i,j)$ by $m_{\mathrm{TSP}}(j)$ as follows:
\begin{equation}
  d^{\mathrm{TSP}}_{\mathrm{mod}}(i,j) = d^{\mathrm{TSP}}_{\mathrm{geo}}(i,j)\, m_{\mathrm{TSP}}(j)
  \label{eq:tsp_mod_cost}
\end{equation}
The scale factor $m_{\mathrm{TSP}}(j)$ is a function of~$s_R^j$, the cosine similarity between the text query and the pooled cell-center embedding~$\mathbf{e}_R$ at~$j$, normalized by the same running maximum of patch--query similarity.
We use the following piecewise map at destination~$j$:
\begin{equation}
  m_{\mathrm{TSP}}(j)=
  \begin{cases}
    1 - 0.8\,\dfrac{s_R^j-0.5}{0.5} & s_R^j>0.5 \\[0.4em]
    1 & s_R^j=0.5 \\[0.4em]
    1 + 5\left(\dfrac{0.5-s_R^j}{1.5}\right)^2 & s_R^j<0.5,
  \end{cases}
  \label{eq:tsp_scale_factor}
\end{equation}
Since the scale factors are destination-based, rescaled costs are generally asymmetric; for example, $d^{\mathrm{TSP}}_{\mathrm{mod}}(i,j)\neq d^{\mathrm{TSP}}_{\mathrm{mod}}(j,i)$ and likewise $d^{\mathrm{SOP}}_{\mathrm{mod}}(i,j)\neq d^{\mathrm{SOP}}_{\mathrm{mod}}(j,i)$. The TSP subproblem over cell centers is an instance of the Asymmetric Traveling Salesman Problem (ATSP), which is solved, as in FALCON, with a Lin--Kernighan--Helsgaun (LKH)-style heuristic~\cite{helsgaun2000effective}.

For $s_F\in[-1,1]$ and $c_F\in[0,1]$, the SOP scale factor~$m_{\mathrm{SOP}}$ lies in $[0.2,1.89]$. Similarly, for~$s_R\in[-1,1]$, the TSP scale factor~$m_{\mathrm{TSP}}$ lies in $[0.2,6]$.
We set the upper end of the SOP scale factor's range below the ceiling for the TSP scale factor because in simulations, raising the SOP bound caused oscillation in the drone's motion.
These bounded, non-zero ranges keep TSP and SOP costs from vanishing or diverging. In summary, semantic rescaling can change the order in which regions or frontiers are explored, but it does not prevent the planners from eventually covering what FALCON would cover.

\section{Experimental Results}
\label{sec:experiments}
This section evaluates language-conditioned exploration with SAGE against geometric, semantic-only, and object-centric baselines on shared indoor scenarios and metrics.

\subsection{Setup and metrics}
We evaluate SAGE in the Exploration Planner Evaluation Environment (EPEE)~\cite{zhang2024falcon} using Matterport3D~\cite{Matterport3D} meshes as indoor environments.
The simulated robot is a quadrotor with a 640$\times$480 RGB--D camera, 0.1~m voxel resolution, and 2--10~Hz replanning. Its dynamics, camera intrinsics, and maximum speeds match the FALCON evaluation setup so that differences in performance can be attributed to the planning algorithm rather than to different vehicle models.

We select five Matterport3D scenes and define two open-vocabulary queries per scene, for a total of ten map--query scenarios.
We run 30 trials for each scenario with different starting positions and headings, and all tables report means over these trials.
FALCON is query-agnostic and does not use language input, so the same 30 trajectories per map are reused for both queries when comparing to SAGE.

We compare SAGE against three planners.
The first one, FALCON, is the geometric baseline that uses the two-level TSP/SOP stack without any semantic information.
The second one, Semantics, is an ablation of SAGE that uses only language for ordering, with TSP and SOP edge costs set to $1{-}s$, where $s$ is the CLIP cosine similarity between the text query and the pooled embedding at the corresponding cell center or frontier, respectively.
Within TSP/SOP for the Semantics baseline, travel distance is omitted from these costs, which removes the geometric coupling in the cost matrices.
The third exploration pipeline, Finding Things in the Unknown (FTU)~\cite{Papatheodorou_ICRA2023}, is an object-centric baseline evaluated in Habitat on the same building layouts.
FTU employs object-centric behavior using instance segmentation and semantic submaps, constraining it to a closed detector vocabulary. As such, only categories that the segmenter is trained to name are available as targets, so it cannot run every open-ended text query we evaluate with CLIP.
Each method uses its native simulator and perception stack.

An object is declared to have been reached when the robot comes within a distance threshold of the ground-truth object's bounding box; we report both 1~m and 1.5~m proximity thresholds.

\subsection{Comparison across planners}
\begin{figure*}[!tp]
  \centering
  \begin{minipage}[t]{0.31\textwidth}\vspace{0pt}
    \centering
    \includegraphics[width=\linewidth,keepaspectratio]{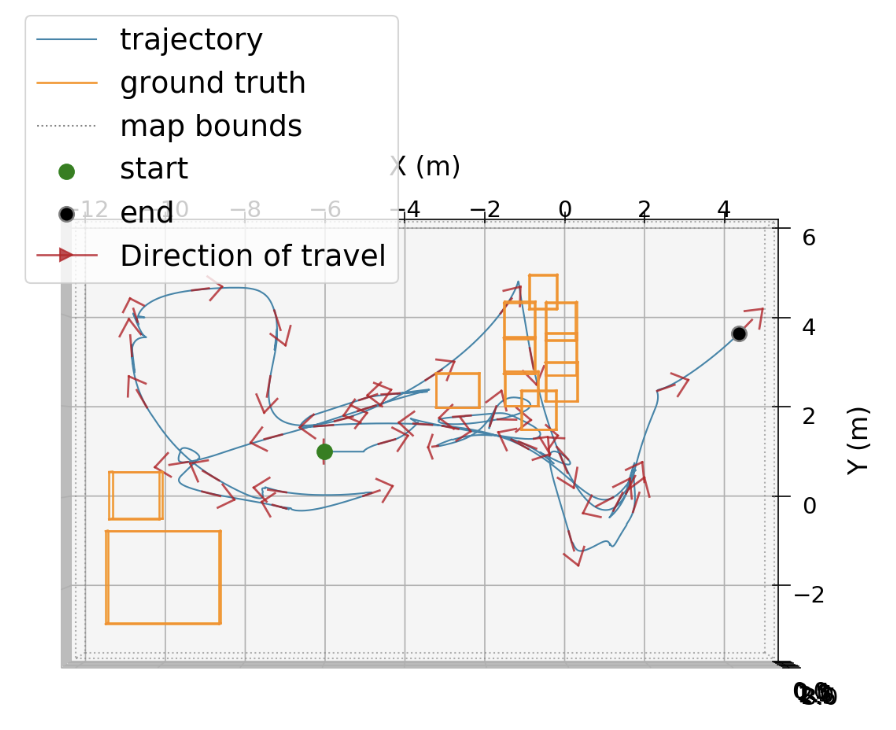}
    \vspace{1mm}
    {\small (a) SAGE}
  \end{minipage}\hfill
  \begin{minipage}[t]{0.31\textwidth}\vspace{0pt}
    \centering
    \includegraphics[width=\linewidth,keepaspectratio]{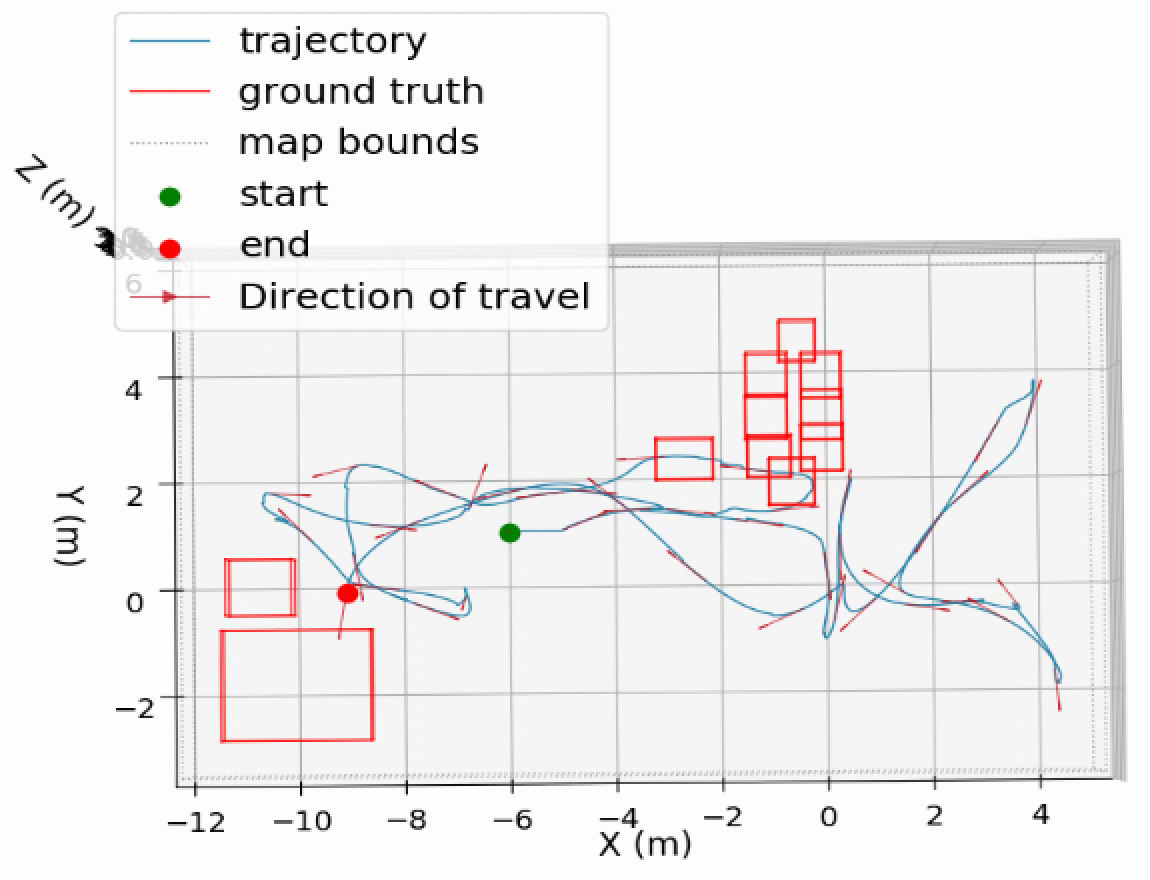}
    \vspace{1mm}
    {\small (b) FALCON}
  \end{minipage}\hfill
  \begin{minipage}[t]{0.31\textwidth}\vspace{0pt}
    \centering
    \includegraphics[width=\linewidth,keepaspectratio]{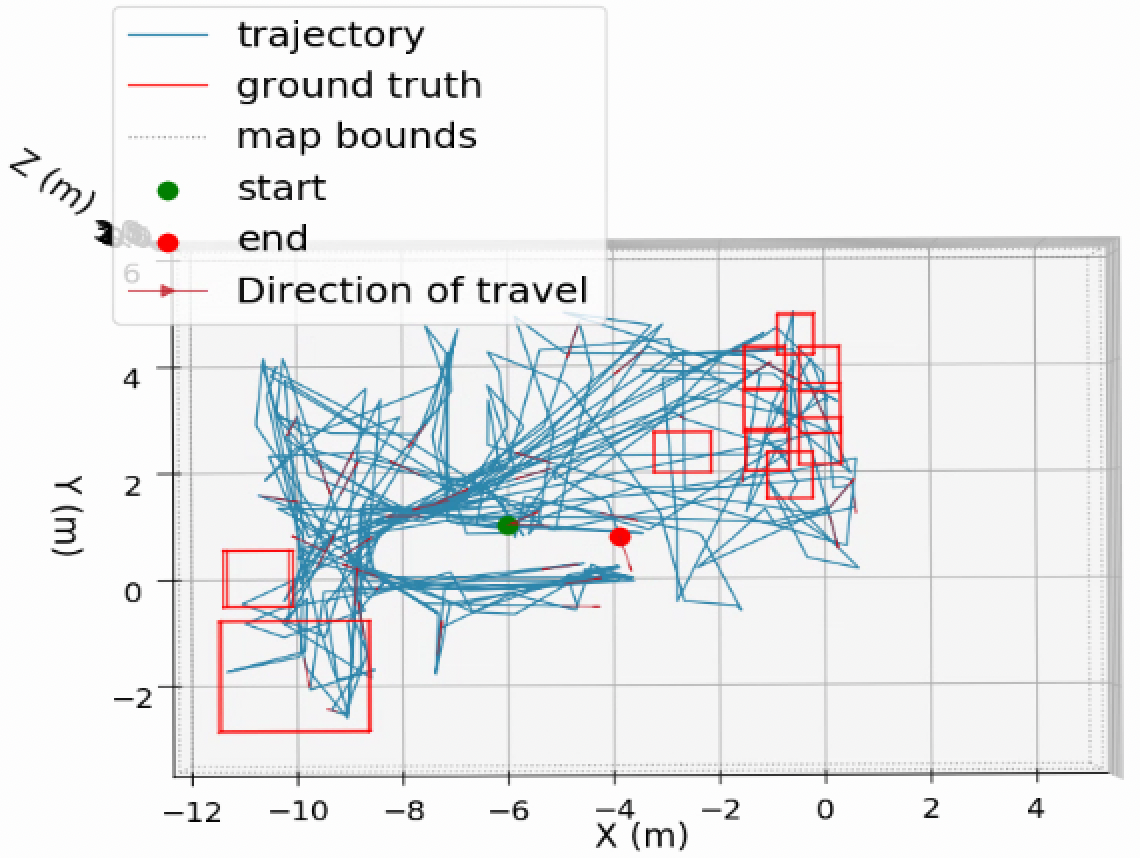}
    \vspace{1mm}
    {\small (c) FTU}
  \end{minipage}
  \caption{Top-down representative trajectory comparison on Map~1, \texttt{chair} query.}
  \label{fig:traj_compare_map1_chair}
\end{figure*}

\begin{figure}[t]
  \centering
  \includegraphics[width=\linewidth]{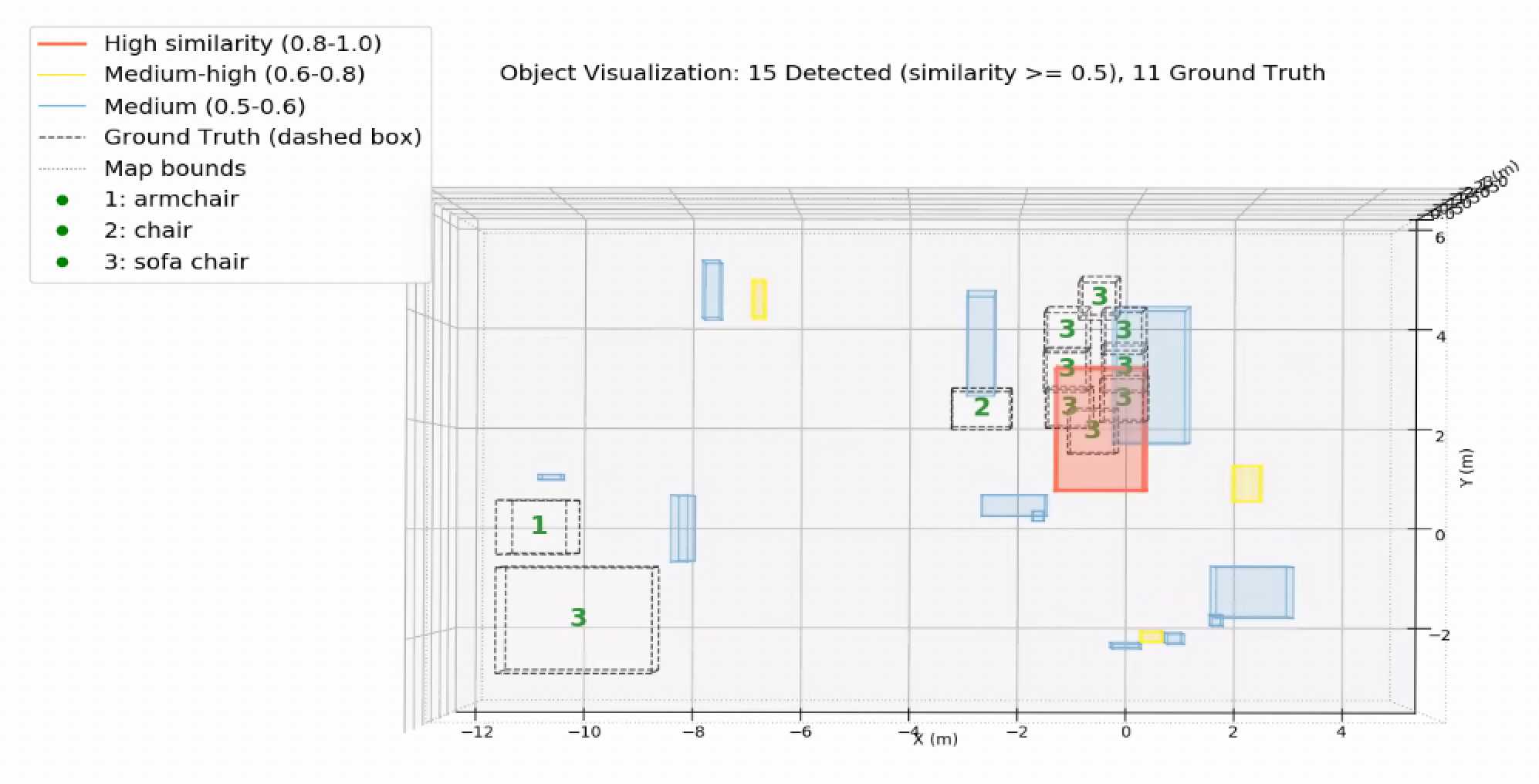}
  \caption{Top-down view of objects with CLIP similarity $\geq 0.5$ vs.\ ground-truth boxes for Map~1, \texttt{chair} query.}
  \label{fig:qual}
\end{figure}

Table~\ref{tab:full} reports performance for all ten scenarios at 1~m and 1.5~m proximity thresholds.
For each planner, we report explored volume, denoted by Vol., total exploration duration, denoted by Dur., path length, denoted by Path, and path length normalized by FALCON's mean path length on the same map--query pair, denoted by $P/P_{\mathrm{F}}$.
Separately for 1~m and 1.5~m proximity, we also report time to first relevant object, denoted by $T_{\mathrm{1st}}$, the ratio of reached to total relevant instances, denoted by R/T, and the number of trials, out of 30, in which all relevant instances are reached, denoted by All$^{\dagger}$.
For each baseline, Table~\ref{tab:sage_reach_ratios} shows the mean, over the map--query pairs in Table~\ref{tab:full}, of the SAGE--to--baseline ratio for duration, path length, and explored volume. For $T_{\mathrm{1st}}$, All$^{\dagger}$, and R/T, the table reports strict win counts at 1~m and 1.5~m, excluding pairs with a tie or no value.\newline
\hspace*{\parindent}As seen in Table~\ref{tab:sage_reach_ratios}, SAGE outperforms FALCON on the semantic metrics $T_{\mathrm{1st}}$, All$^{\dagger}$, and R/T.
Conversely, FALCON outperforms SAGE on the purely geometric metrics of duration, path length, and explored volume.
Both of these are to be expected, since FALCON is purely geometric and does not account for semantics.\newline
\hspace*{\parindent}From Table~\ref{tab:sage_reach_ratios}, we also see that SAGE outperforms Semantics on All$^{\dagger}$ and R/T at both proximity thresholds, while the two are on par on $T_{\mathrm{1st}}$.
As for geometric metrics, SAGE outperforms Semantics on path and duration and is on par with Semantics for volume.
Since the semantic-only ablation omits travel distance in edge costs, minimizing its semantic objective alone sends the robot on long hops toward visually salient regions.
It may also repeatedly visit high-similarity areas while leaving distant unknown regions unexplored until late in the mission.
This is the classic failure mode of pure ``semantic attraction'' without an explicit map-completion drive and motivates SAGE's unified semantic--geometric cost.\newline
\hspace*{\parindent}From Table~\ref{tab:sage_reach_ratios}, SAGE significantly outperforms FTU in terms of geometric metrics such as volume, duration, and path length.
However, FTU outperforms SAGE on All$^{\dagger}$ and R/T.
This is to be expected, since FTU's object-centric pipeline devotes more perception to instance-level object views.
Interestingly, for $T_{\mathrm{1st}}$, SAGE outperforms FTU at both proximity thresholds.
We attribute this advantage, in part, to SAGE's real-time planning and faster exploration, which can bring the robot to query-relevant objects more quickly.
SAGE finishes exploration $9.0$ to $25.9$ times faster than FTU, with a mean speedup of $13.7$.
Exploration efficiency, measured in volume per second, for SAGE is $9.5$ to $80.4$ times higher, with a mean of $25.1$ times relative to FTU.\newline
\begin{table*}[!tp]
  \centering
  \caption{Simulation means over 30 trials. \textbf{Bold}: best value in each column among the four planners in that map--query block; ties are both bold. FTU values use the native Habitat stack; Map~5 \texttt{shelf} is unsupported in FTU and shown as ``--''. In the $T_{\mathrm{1st}}$ column under 1.5~m proximity, ``--'' indicates the target is already within 1.5~m at mission start.}
  \label{tab:full}
  \adjustbox{width=\linewidth,max height=0.42\textheight,center}{%
  \fontsize{5.1pt}{5.72pt}\selectfont
  \setlength{\tabcolsep}{2.55pt}
  \renewcommand{\arraystretch}{0.845}
  \setlength{\aboverulesep}{0.35ex}
  \setlength{\belowrulesep}{0.4ex}
  \begin{tabular}{@{}cllrrrr|rrr|rrr@{}}
    \toprule
    \multirow{2}{*}{Map} & \multirow{2}{*}{Query} & \multirow{2}{*}{Method} & \multirow{2}{*}{Dur.$\downarrow$} & \multirow{2}{*}{Path (m)$\downarrow$} & \multirow{2}{*}{$P/P_{\mathrm{F}}\downarrow$} & \multirow{2}{*}{Vol.$\uparrow$} & \multicolumn{3}{c|}{\textbf{1~m proximity}} & \multicolumn{3}{c}{\textbf{1.5~m proximity}} \\
    \multicolumn{7}{c|}{} & $T_{\mathrm{1st}}\downarrow$ & R/T$\uparrow$ & All$^{\dagger}\uparrow$ & $T_{\mathrm{1st}}\downarrow$ & R/T$\uparrow$ & All$^{\dagger}\uparrow$ \\
    \cmidrule(lr){8-10}\cmidrule(lr){11-13}
    \midrule
    \multirow{8}{*}{1} & \multirow{4}{*}{\texttt{chair}} & SAGE & 72.05 & 92.78 & 1.22 & 267.07 & 6.02 & 8.3/11 & 5 & 4.04 & 9.8/11 & 11 \\
     &                        & FALCON & \textbf{52.93} & \textbf{76.26} & \textbf{1.00} & \textbf{268.66} & 9.95 & 6.9/11 & 0 & 7.56 & 9.0/11 & 8 \\
     &                        & Semantics & 74.69 & 98.68 & 1.29 & 266.35 & \textbf{3.81} & 7.4/11 & 0 & \textbf{2.88} & 9.2/11 & 8 \\
    &                        & FTU & 659.88 & 541.76 & 7.10 & 222.85 & 8.51 & \textbf{10.9/11} & \textbf{28} & 5.20 & \textbf{11/11} & \textbf{30} \\
     \cmidrule{2-13}
     & \multirow{4}{*}{\texttt{bed}} & SAGE & 70.20 & 95.69 & 1.25 & 267.50 & 42.71 & 2.6/4 & 7 & 41.16 & 3.3/4 & 11 \\
    &                        & FALCON & \textbf{52.93} & \textbf{76.26} & \textbf{1.00} & \textbf{268.66} & \textbf{25.44} & 2.6/4 & 8 & 24.52 & 3.4/4 & 12 \\
    &                        & Semantics & 70.57 & 100.93 & 1.32 & 264.76 & 41.02 & 2.5/4 & 5 & 40.28 & 3.2/4 & 10 \\
   &                        & FTU & 632.63 & 552.20 & 7.24 & 229.26 & 35.21 & \textbf{4/4} & \textbf{30} & \textbf{8.83} & \textbf{4/4} & \textbf{30} \\
    \midrule
    \multirow{8}{*}{2} & \multirow{4}{*}{\texttt{chair}} & SAGE & 73.71 & 90.14 & 1.08 & 305.77 & 11.85 & 11.1/15 & 1 & \textbf{9.89} & 13.0/15 & 6 \\
     &                        & FALCON & \textbf{60.38} & \textbf{83.71} & \textbf{1.00} & \textbf{323.47} & 14.93 & 9.9/15 & 0 & 11.63 & 12.8/15 & 4 \\
     &                        & Semantics & 78.75 & 103.33 & 1.23 & 299.70 & \textbf{11.28} & 11.6/15 & 1 & 10.95 & 13.5/15 & 9 \\
    &                        & FTU & 1243.63 & 711.58 & 8.50 & 296.23 & 18.42 & \textbf{15/15} & \textbf{30} & 12.05 & \textbf{15/15} & \textbf{30} \\
     \cmidrule{2-13}
     & \multirow{4}{*}{\texttt{picture}} & SAGE & 72.86 & 86.23 & 1.03 & 297.30 & \textbf{3.53} & 13.5/20 & 0 & -- & 16.9/20 & 0 \\
    &                        & FALCON & \textbf{60.38} & \textbf{83.71} & \textbf{1.00} & \textbf{323.47} & 6.73 & 10.8/20 & 0 & -- & 15.3/20 & 0 \\
    &                        & Semantics & 77.61 & 104.75 & 1.25 & 299.52 & 5.19 & 12.9/20 & 0 & -- & \textbf{17.3/20} & 2 \\
   &                        & FTU & 658.15 & 382.39 & 4.57 & 266.20 & 32.90 & \textbf{19.4/20} & \textbf{12} & -- & \textbf{19.8/20} & \textbf{23} \\
    \midrule
    \multirow{8}{*}{3} & \multirow{4}{*}{\texttt{chair}} & SAGE & \textbf{14.09} & 17.91 & 1.12 & 117.63 & 4.68 & 2.9/5 & \textbf{0} & \textbf{2.20} & 4.2/5 & \textbf{12} \\
     &                        & FALCON & 15.32 & \textbf{15.93} & \textbf{1.00} & \textbf{128.09} & \textbf{3.59} & 2.6/5 & \textbf{0} & 2.51 & 4.0/5 & 9 \\
     &                        & Semantics & 15.75 & 19.37 & 1.22 & 118.30 & 3.75 & 2.8/5 & \textbf{0} & 2.32 & \textbf{4.3/5} & 11 \\
     &                        & FTU & 258.54 & 101.09 & 6.35 & 36.76 & 19.11 & \textbf{3.9/5} & \textbf{0} & 14.21 & 4.0/5 & 0 \\
     \cmidrule{2-13}
     & \multirow{4}{*}{\texttt{table}} & SAGE & \textbf{14.70} & 18.50 & 1.16 & 118.39 & 4.16 & 4.1/8 & \textbf{0} & 3.03 & \textbf{6.9/8} & \textbf{8} \\
   &                        & FALCON & 15.32 & \textbf{15.93} & \textbf{1.00} & \textbf{128.09} & \textbf{3.17} & 3.4/8 & \textbf{0} & \textbf{2.22} & 5.9/8 & 1 \\
   &                        & Semantics & 15.64 & 19.07 & 1.20 & 114.50 & 5.38 & \textbf{4.2/8} & \textbf{0} & 4.65 & 6.8/8 & 0 \\
   &                        & FTU & 381.15 & 139.40 & 8.75 & 38.20 & 22.98 & 1.6/8 & \textbf{0} & 19.41 & 1.7/8 & 0 \\
    \midrule
    \multirow{8}{*}{4} & \multirow{4}{*}{\texttt{chair}} & SAGE & 35.73 & \textbf{40.75} & \textbf{0.97} & 115.71 & \textbf{2.31} & 7.8/9 & 9 & -- & 8.6/9 & 20 \\
     &                        & FALCON & \textbf{25.76} & 41.94 & 1.00 & \textbf{133.75} & 5.28 & 6.2/9 & 0 & -- & 8.0/9 & 11 \\
     &                        & Semantics & 41.78 & 42.35 & 1.01 & 112.59 & 2.39 & 7.2/9 & 4 & -- & 8.5/9 & 20 \\
    &                        & FTU & 492.82 & 315.77 & 7.53 & 105.99 & 14.21 & \textbf{9/9} & \textbf{30} & -- & \textbf{9/9} & \textbf{30} \\
     \cmidrule{2-13}
     & \multirow{4}{*}{\texttt{mirror}} & SAGE & 37.42 & \textbf{37.39} & \textbf{0.89} & 112.83 & \textbf{2.76} & 1.8/3 & 4 & -- & \textbf{2.6/3} & 18 \\
    &                        & FALCON & \textbf{25.76} & 41.94 & 1.00 & \textbf{133.75} & 2.98 & 1.5/3 & 0 & -- & 1.9/3 & 7 \\
    &                        & Semantics & 39.87 & 40.98 & 0.98 & 113.62 & 3.23 & 1.6/3 & 0 & -- & 2.2/3 & 9 \\
   &                        & FTU & 353.93 & 223.26 & 5.32 & 112.22 & 9.21 & \textbf{2.0/3} & \textbf{5} & -- & \textbf{2.6/3} & \textbf{21} \\
    \midrule
    \multirow{8}{*}{5} & \multirow{4}{*}{\texttt{chair}} & SAGE & 64.46 & 78.32 & 1.10 & 239.89 & 11.45 & \textbf{9.4/12} & \textbf{0} & 10.73 & 11.5/12 & 16 \\
     &                        & FALCON & \textbf{51.32} & \textbf{71.01} & \textbf{1.00} & \textbf{246.87} & 15.81 & 8.4/12 & \textbf{0} & 14.13 & 11.1/12 & 12 \\
     &                        & Semantics & 65.19 & 87.72 & 1.24 & 235.48 & \textbf{10.67} & 8.9/12 & \textbf{0} & \textbf{9.68} & \textbf{11.6/12} & \textbf{18} \\
     &                        & FTU & 756.13 & 527.83 & 7.43 & 207.18 & 22.78 & 9.0/12 & \textbf{0} & 21.92 & 10.9/12 & 2 \\
     \cmidrule{2-13}
     & \multirow{4}{*}{\texttt{shelf}} & SAGE & 62.35 & 76.43 & 1.08 & 241.04 & 9.86 & 8.6/13 & \textbf{0} & -- & \textbf{11.7/13} & \textbf{6} \\
   &                        & FALCON & \textbf{51.32} & \textbf{71.01} & \textbf{1.00} & \textbf{246.87} & 11.80 & 7.5/13 & \textbf{0} & -- & 11.1/13 & 4 \\
   &                        & Semantics & 64.75 & 86.11 & 1.21 & 239.32 & \textbf{8.94} & \textbf{9.0/13} & \textbf{0} & -- & 11.3/13 & 1 \\
    &                        & FTU & -- & -- & -- & -- & -- & -- & -- & -- & -- & -- \\
    \bottomrule
  \end{tabular}}
\end{table*}
\begin{table}[!tp]
  \centering
  \caption{Pairwise comparison of SAGE vs other baselines in simulation, obtained by averaging and aggregating results from Table~\ref{tab:full}.}
  \label{tab:sage_reach_ratios}
  {\setlength{\tabcolsep}{2.6pt}
  \setlength{\aboverulesep}{0.25ex}
  \setlength{\belowrulesep}{0.35ex}
  \renewcommand{\arraystretch}{1.02}
  \small
  \begin{tabular*}{\columnwidth}{@{\extracolsep{\fill}}@{}c l@{\hspace{0.35em}}c c c|c c c@{}}
    \toprule
    \makecell{\textbf{Thr.}} & \makecell{\textbf{Baseline}} & \textbf{$T_{\mathrm{1st}}\uparrow$} & \textbf{All$^{\dagger}\uparrow$} & \textbf{R/T$\uparrow$} & \textbf{Dur.$\downarrow$} & \textbf{Path$\downarrow$} & \textbf{Vol.$\uparrow$} \\
    \cmidrule(lr){3-5}
    \cmidrule(lr){6-8}
    \midrule
    \multirow{3}{*}{1~m} & FTU & 8/9 & 0/6 & 2/9 & 0.08 & 0.16 & 1.56 \\
    & FALCON & 7/10 & 4/5 & 9/9 & 1.23 & 1.09 & 0.94 \\
    & Sem. & 4/10 & 4/4 & 7/10 & 0.94 & 0.91 & 1.01 \\
    \midrule
    \multirow{3}{*}{1.5~m} & FTU & 5/6 & 3/9 & 3/8 & 0.08 & 0.16 & 1.56 \\
    & FALCON & 4/6 & 8/9 & 9/10 & 1.23 & 1.09 & 0.94 \\
    & Sem. & 3/6 & 6/9 & 6/10 & 0.94 & 0.91 & 1.01 \\
    \bottomrule
  \end{tabular*}%
  }%
\end{table}
\hspace*{\parindent}Figure~\ref{fig:traj_compare_map1_chair} compares representative trajectories for SAGE, FALCON, and FTU on the same \texttt{chair} query in Map~1.
FALCON follows a shorter geometry-driven sweep that emphasizes space coverage, while FTU spends longer around target-bearing regions to obtain repeated object viewpoints.
SAGE lies between these extremes, as it maintains broad map progression while still biasing motion toward semantically relevant regions.
This behavior matches Table~\ref{tab:full} on this \texttt{chair} query in that SAGE attains higher mean reached counts than FALCON while finishing far sooner than FTU on the same layout.\newline
\hspace*{\parindent}Figure~\ref{fig:qual} shows detections for a \texttt{chair} query on Map~1 for SAGE, where high-similarity objects cluster near annotated chairs, validating that CLIP scores provide meaningful signals for the planner.

\subsection{Hardware Experiments}

\begin{figure}[!htbp]
  \centering
  \begin{minipage}[t]{0.50\columnwidth}\vspace{0pt}
    \centering
    \includegraphics[width=\linewidth,keepaspectratio]{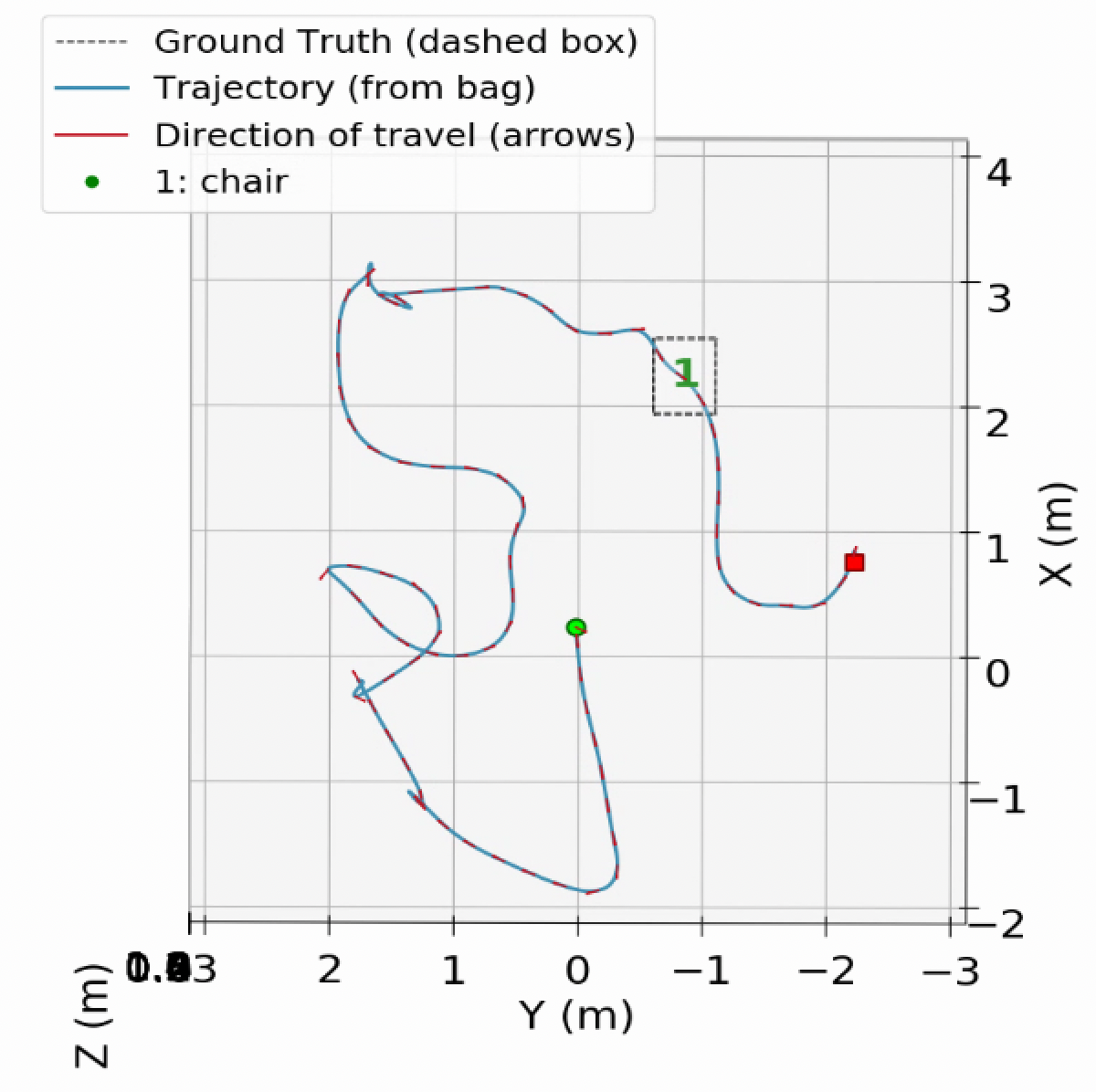}
    \vspace{1mm}
    {\small (a) SAGE, Trial~4 \texttt{chair}}
  \end{minipage}\hfill
  \begin{minipage}[t]{0.50\columnwidth}\vspace{0pt}
    \centering
    \includegraphics[width=\linewidth,keepaspectratio]{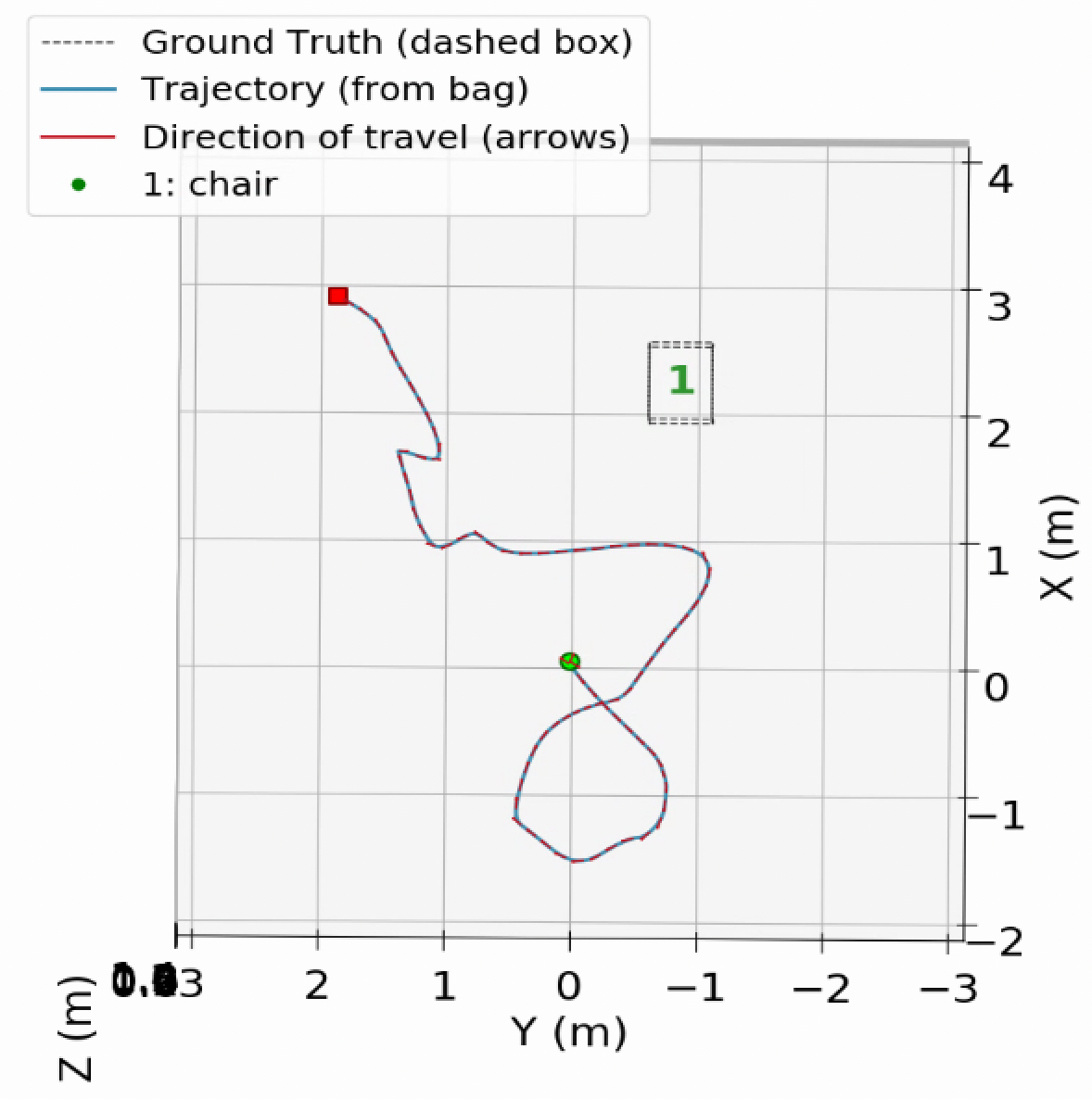}
    \vspace{1mm}
    {\small (b) FALCON, same start as SAGE, Trial~4}
  \end{minipage}
  \caption{Top-down hardware flight trajectories with approximate ground-truth reference. SAGE Trial~4 biases motion toward the queried chair; FALCON follows a shorter geometric exploration path under the same start pose, cf.\ Table~\protect\ref{tab:trial4_matched}.}
  \label{fig:hw_traj_sage_falcon}
\end{figure}

We deploy SAGE on the Modal AI Starling~2 quadrotor with a time-of-flight depth sensor, a high-resolution RGB camera, and visual--inertial odometry~\cite{Geneva2020ICRA}, reusing the same ROS nodes as in simulation.
For hardware trials, onboard compute limitations of the Qualcomm QRB5165 processor force us to offload CLIP inference to a workstation with an NVIDIA GeForce RTX~3080 on the same Wi-Fi network.
The drone publishes RGB images over ROS; the workstation runs CLIP inference and returns the resulting 512-D embedding in the service response for the drone planner to use.
We ran five indoor SAGE flights in two environments. Trials~1--3 were in one environment, with two \texttt{chair} targets, one \texttt{table} target, and one \texttt{backpack} target. Trials~4--5 were in a second environment, with one \texttt{chair} target per trial and different initial headings relative to the chair.
In Trials~1--3, each incoming depth frame was paired with an RGB image for CLIP inference.
In Trials~4--5, we increased temporal frame downsampling by a factor of 5 in the RGB stream. At a 30~Hz RGB frame rate, this corresponds to one CLIP update every 0.5~s, which aligns with every fifth depth frame in a 10~Hz stream. This reduces contention between semantic inference and map-update processing and lowers effective map-update latency during flight.
Other pipeline settings were unchanged.\newline
\hspace*{\parindent}Figure~\ref{fig:hw_traj_sage_falcon} compares SAGE and FALCON under identical Trial~4 start conditions in the second environment.
SAGE reaches the 1~m threshold for the queried chair, while FALCON stays farther from the chair as it prioritizes geometric expansion.
As seen in Table~\ref{tab:trial4_matched} for Trial~4, coverage is 98.7\% for SAGE and 99.0\% for FALCON, indicating that both planners achieve near-complete coverage of the same task box. In the same comparison, SAGE and FALCON path lengths are 25.08~m and 16.82~m, respectively. This demonstrates FALCON's advantage in exploration efficiency as expected.
Table~\ref{tab:hardware} summarizes per-trial reach and timing for all five SAGE flights at 1~m and 1.5~m thresholds.
Across all five flights, SAGE reaches every target instance at both thresholds.
Coverage remains high in both environments, averaging 91.7\% and 95.8\% of the predefined 49.2~m$^3$ and 52.5~m$^3$ task-box volumes in the first and second environments, respectively.

\begin{table}[!tbp]
  \centering
  \caption{Hardware experiments for trial~4 in Table~\ref{tab:hardware} (matched SAGE/FALCON start; thresholds 1~m and 1.5~m).}
  \label{tab:trial4_matched}
  {\setlength{\tabcolsep}{2.8pt}
  \renewcommand{\arraystretch}{1.02}
  \small
  \begin{tabular*}{\columnwidth}{@{\extracolsep{\fill}}@{}l c c c c c c@{}}
    \toprule
    \textbf{Meth.} & \makecell{\textbf{1m R/T}\\[-1.5pt] $\uparrow$} & \makecell{\textbf{1.5m R/T}\\[-1.5pt] $\uparrow$} & \makecell{\textbf{Dur.}\\(s) $\downarrow$} & \makecell{\textbf{Path}\\(m) $\downarrow$} & \makecell{\textbf{Vol.}\\(m$^3$) $\uparrow$} & \makecell{\textbf{Cov.}\\(\%) $\uparrow$} \\
    \midrule
    SAGE & \textbf{1/1} & \textbf{1/1} & 29.36 & 25.08 & 51.81 & 98.7 \\
    FALCON & 0/1 & \textbf{1/1} & \textbf{14.86} & \textbf{16.82} & \textbf{51.97} & \textbf{99.0} \\
    \bottomrule
  \end{tabular*}%
  }%
\end{table}

\begin{table}[!htbp]
  \centering
  \caption{Hardware deployment metrics on the Modal AI Starling~2 at both proximity thresholds.}
  \label{tab:hardware}
  {\setlength{\tabcolsep}{1.7pt}
  \renewcommand{\arraystretch}{1.02}
  \small
  \begin{tabular*}{\columnwidth}{@{\extracolsep{\fill}}@{}c c l@{\hspace{0.2em}}c c@{\hspace{0.12em}}c c r r r@{}}
    \toprule
    \multirow{2}{*}{\makecell{\textbf{Env.}}} & \multirow{2}{*}{\textbf{Trial}} & \multirow{2}{*}{\makecell{\textbf{Query}}} & \multicolumn{2}{c}{\textbf{1m}} & \multicolumn{2}{c}{\textbf{1.5m}} & \multirow{2}{*}{\makecell{\textbf{Dur.}\\(s)$\downarrow$}} & \multirow{2}{*}{\makecell{\textbf{Vol.}\\(m$^3$)$\uparrow$}} & \multirow{2}{*}{\makecell{\textbf{Cov.}\\\%$\uparrow$}} \\
    & & & R/T$\uparrow$ & $T_{\mathrm{1st}}\downarrow$ & R/T$\uparrow$ & $T_{\mathrm{1st}}\downarrow$ & & & \\
    \cmidrule(lr){4-5}\cmidrule(lr){6-7}
    \midrule
    \multirow{3}{*}{1} & 1 & chair & 2/2 & 10.3 & 2/2 & 9.7 & 71.6 & 45.8 & 93.0 \\
    & 2 & table & 1/1 & 43.6 & 1/1 & 16.5 & 54.1 & 45.5 & 92.5 \\
    & 3 & backpack & 1/1 & 21.1 & 1/1 & 20.5 & 50.4 & 44.2 & 89.7 \\
    \cmidrule{1-10}
    \multirow{2}{*}{2} & 4 & chair & 1/1 & 14.7 & 1/1 & 13.8 & 29.4 & 51.8 & 98.7 \\
    & 5 & chair & 1/1 & 2.5 & 1/1 & 1.9 & 38.8 & 48.8 & 92.9 \\
    \midrule
    \multicolumn{7}{r}{\textit{Mean (all 5 trials)}} & 48.9 & 47.2 & 93.4 \\
    \bottomrule
  \end{tabular*}%
  }%
  \vspace{1mm}
\end{table}

\section{Conclusions}
\label{sec:conclusion}

We propose a semantic-aware exploration system, SAGE, that couples open-vocabulary CLIP similarity with FALCON's volumetric exploration through semantic--geometric costs, compact object-based memory, a temporal cache for frontier semantics, and object frontier viewpoints.
Across Matterport3D simulations, SAGE outperforms geometric and semantic-only baselines in object discovery and is more than an order of magnitude faster than FTU-style object-centric exploration in matched layouts, with hardware validation on the Starling~2 platform.

Visual--inertial odometry drift and vehicle dynamics on real MAVs remain dominant deployment risks.
As visual--inertial odometry quality degrades, mapped object instances are located at incorrect map locations relative to the drone, so semantic memory can become spatially skewed and objects in semantic memory may no longer align with the true target positions.
In our current hardware pipeline, semantics are assigned from one CLIP embedding per time-of-flight frame rather than patch-level embeddings, which simplifies integration but reduces spatial semantic precision.
The mission-relative running maximum of patch--query CLIP similarity normalizes semantic scores during each mission, improving robustness to varying RGB appearance across our evaluated indoor environments.

Future work includes stronger multi-view fusion and temporally consistent object tracks that could stabilize embeddings under motion blur and other sensing artifacts.
Techniques such as model distillation, smaller vision transformers, and neural-processing-unit-aware training could yield onboard-friendly models that reduce dependence on offboard CLIP inference.
Extending costs to dynamic scenes, multi-agent semantic maps, and tighter coupling with LLM-decomposed task graphs are natural next steps for 3D-LLM and vision--language--action settings where language specifies subgoals over hours-long missions.

\clearpage
{
    \small
    \bibliographystyle{ieeenat_fullname}
    \bibliography{references}
}

\end{document}